\documentclass{article}

\usepackage{arxiv}

\usepackage[utf8]{inputenc} 
\usepackage[T1]{fontenc}    
\usepackage{hyperref}       
\usepackage{url}            
\usepackage{booktabs}       
\usepackage{amsfonts}       
\usepackage{nicefrac}       
\usepackage{microtype}      
\usepackage{lipsum}
\usepackage{times}
\usepackage{epsfig}
\usepackage{graphicx}
\usepackage{amsmath}
\usepackage{amssymb}
\usepackage{rotating}
\usepackage{multirow}
\usepackage{subfig}
\usepackage{bm}
\usepackage{hhline}
\usepackage[norule,symbol,perpage]{footmisc}
\title{A Valid Watermark Is Not a Valid Identity: Spatial-Spectral Adversarial Face Evasion with Watermark Preservation}

\author{
Jefferson David Rodriguez Chivata, Davide Ghiani, Simone Maurizio La Cava,\\
Marco Micheletto, Giulia Orr\`u, Gian Luca Marcialis\\
University of Cagliari, Piazza d'Armi I, 09123 Cagliari, Italy\\
\texttt{\small jeffersond.rodriguez, davide.ghiani, simonem.lac,}\\
\texttt{\small marco.micheletto, giulia.orr\`u, marcialis@unica.it}
}

\usepackage{tikz}
\usepackage{textcomp}
\usepackage{hyperref}

\newcommand{\copyrighttext}{%
\footnotesize
\textcopyright\ 2026 IEEE. Personal use of this material is permitted.
Permission from IEEE must be obtained for all other uses, in any current
or future media, including reprinting/republishing this material for
advertising or promotional purposes, creating new collective works,
for resale or redistribution to servers or lists, or reuse of any
copyrighted component of this work in other works. DOI:
}

\newcommand{\copyrightnotice}{%
\begin{tikzpicture}[remember picture,overlay]
\node[anchor=south,yshift=10pt] at (current page.south)
{\parbox{\textwidth}{\copyrighttext}};
\end{tikzpicture}%
}
\begin{document}
\maketitle
\copyrightnotice
\begin{abstract}
   Facial image watermarking can support traceability and certification in biometric pipelines, but provenance or manipulation detection alone does not guarantee reliable biometric identity. We introduce a watermark-preserving face evasion attack that targets this gap under a black-box watermarking setting, without access to the watermarking components. The proposed PGD-$L_2$ attack guides adversarial perturbations with spatial and spectral preservation terms to reduce face-recognition similarity while limiting pixel-level distortion and high-frequency changes that may interfere with hidden watermark cues. Evaluated on controlled facial images with two watermarking schemes, two face-recognition models, and two datasets, the attack achieves evasion rates above $92\%$ at operational thresholds while preserving watermark recovery above $93\%$ for StegaFormer and above $99.7\%$ for FaceSigns, with visual quality above $47$~dB PSNR at $\varepsilon_2=2.0$. These results expose a structural gap in watermark-based facial certification: a face image can remain watermark-valid and visually faithful while becoming biometrically unreliable.
\end{abstract}

\section{Introduction}

Face images used in identity documents, border control, and remote verification systems must satisfy two distinct security requirements: they must be traceable to verify provenance, the issuing authority, or subsequent alteration, and they must remain biometrically reliable to support automated identity verification. Standards such as ICAO Doc 9303 \cite{icao20159303} and ISO/IEC 39794-5:2019 \cite{ISO39794_5} define technical requirements for face images in identity contexts, while regulatory and provenance frameworks, including the EU Artificial Intelligence Act \cite{eu_ai_act_2024} and C2PA-related initiatives \cite{england2021amp}, reinforce the need for auditable visual evidence. Cryptographic signatures bind provenance to an exact byte sequence and do not survive the re-encoding and resizing typical of issuance pipelines, which is why provenance frameworks complement them with pixel-level soft bindings such as invisible watermarking \cite{chivata2025deep}. 
Recent neural watermarking methods improve the trade-off among imperceptibility, robustness, payload, and recovery accuracy~\cite{tancik2020stegastamp,zhu2018hidden}. In facial media, watermarking has also been extended to authentication and proactive forensics through semi-fragile designs~\cite{nadimpalli2024social,neekhara2024facesigns, ghiani2025fragile}, tracing/manipulation-detection ~\cite{liu2023bifpro,wu2023sepmark}, and robustness evaluation benchmarks~\cite{an2024waves,lu2025robust}. However, these evaluations mainly assess watermark survival or manipulation detection, rather than verifying if a watermarked face remains reliable for biometric recognition.

A critical gap arises when watermark validity is implicitly treated as a proxy for image trustworthiness. A correctly recovered watermark can certify provenance, but it does not necessarily certify that the facial representation remains valid for biometric verification. Prior work has shown that facial biometric evidence can be deliberately degraded after generative editing as a privacy-preserving defense \cite{wang2025edit}. The failure mode studied here is different: an already watermarked facial image is adversarially modified so that the embedded watermark remains recoverable, while the face fails to match its own enrollment template under automated 1:1 verification. The perturbation drives the facial embedding away from its own reference, so that, to a human examiner, the image remains visually the same person. The resulting image is neither a watermark-removal case, in which the decoder fails, nor a watermark-forgery case, in which the decoder succeeds on an unwatermarked image. Instead, the decoder succeeds, and the watermark remains valid, but the face-recognition system rejects the same image.

This distinction is important because semi-fragile facial watermarking schemes are typically designed to react to structural facial manipulations, such as face swapping or deepfake generation, rather than to small adversarial perturbations~\cite{neekhara2024facesigns}. Such perturbations do not replace the face or visibly alter its semantic content. They introduce coordinated pixel-level changes that may appear, from the watermark decoder perspective, similar to benign image degradation. As a result, the image can still be certified as traceable while its biometric content is compromised.

Existing research has studied adversarial attacks against face recognition and watermarking for facial media, but the inverse scenario considered here remains largely unexplored. Attacks such as FGSM~\cite{goodfellow2014explaining} and PGD~\cite{madry2017towards}, including face-specific physical and digital variants~\cite{dong2019efficient,sharif2016accessorize}, show that facial representations can be disrupted. Conversely, recent watermarking works have designed adversarial or dual-purpose watermarks for face swapping, deepfake detection, or identity protection~\cite{wu2024watermarks,yao2024adversarial,zhang2026dip,zhang2024dual}. In those cases, the watermarking process is modified, optimized, or exploited. Here, the adversary receives an already watermarked face image, treats the watermarking system as a black box, and optimizes a perturbation that evades face recognition while preserving the existing watermark.

The present work formulates this inverse scenario as a security evaluation of watermarked biometric pipelines. The threat model reflects a realistic certification setting: in provenance and issuance pipelines the watermarking system belongs to the issuing authority, so a downstream adversary receives an already watermarked image and must treat it as black-box. The face-recognition model is instead assumed white-box, as a conservative worst-case on the biometric side—if the watermark–identity gap emerges against a fully known recognizer, the vulnerability is established, with transferability to unseen recognizers left to future work. Given a watermarked probe image, the attacker optimizes an $L_2$-bounded perturbation restricted to the facial region to reduce cosine similarity with the mated reference template. Because the watermark encoder, decoder, and message are unavailable to the attacker, as they remain under the issuing authority's control, watermark preservation is enforced indirectly through spatial and spectral constraints. The spatial term limits pixel-level deviation from the watermarked image, while the spectral term penalizes high-frequency perturbation energy that may interfere with hidden watermark cues, in line with evidence that adversarial watermarking residuals emphasize high-frequency structures \cite{yao2024adversarial} and that watermark robustness can be affected by suppression of mid- and high-frequency patterns \cite{lu2025robust}.

The paper makes three contributions. First, it formulates watermark-preserving biometric evasion as a security task, with a threat model granting black-box watermark access and white-box recognizer access. Second, it introduces a PGD-$L_2$ attack with spatial and spectral preservation that requires no access to the watermark encoder, decoder, or embedded message. Third, it evaluates the attack across two watermark-capacity regimes, represented by FaceSigns~\cite{neekhara2024facesigns} and StegaFormer~\cite{yu2024effective}, on two controlled face datasets and two face-recognition models. The results show that watermark validity and biometric identity are independent properties: a face image can remain traceable while becoming unreliable for automated verification.

\section{Related Works}
\label{related_works}

Digital watermarking embeds hidden information in images for ownership verification, provenance tracking, and authentication. Neural methods such as HiDDeN~\cite{zhu2018hidden}, StegaStamp~\cite{tancik2020stegastamp}, and MBRS~\cite{jia2021mbrs} formulate watermarking as an end-to-end encoder--decoder problem that jointly optimizes imperceptibility, robustness, payload capacity, and message recovery. More recent high-capacity methods, including StegaFormer~\cite{yu2024effective}, further increase payload density while maintaining high recovery and visual fidelity.

In facial media, watermarking has been adapted to authentication and proactive forensics. FaceSigns~\cite{neekhara2024facesigns} and related semi-fragile schemes~\cite{nadimpalli2024social} preserve the watermark under benign transformations while reacting to facial manipulations. SepMark~\cite{wu2023sepmark} and BiFPro~\cite{liu2023bifpro} separate tracing and manipulation detection through watermark components with different robustness profiles, while benchmarks such as WAVES~\cite{an2024waves} and W-Bench~\cite{lu2025robust} evaluate watermark survival under distortions, regeneration, editing, and adversarial attacks. However, these evaluations mainly assess whether the watermark can still be detected or decoded. They do not test whether the same watermarked image remains reliable for biometric recognition.

Adversarial attacks against face recognition provide the complementary perspective. General attacks such as FGSM~\cite{goodfellow2014explaining} and PGD~\cite{madry2017towards} showed that small norm-bounded perturbations can alter deep-network decisions, while face-specific studies demonstrated this vulnerability in physical and digital verification settings~\cite{dong2019efficient,sharif2016accessorize}. Recent biometric-defense methods further show that adversarial perturbations can deliberately degrade identity representations after generative editing~\cite{wang2025edit}. These works demonstrate that face representations can be selectively disrupted, but they do not consider already-watermarked facial images or measure whether watermark recovery survives the attack.

Several works combine watermarking and adversarial signals for active protection against facial manipulation. CMUA-Watermark~\cite{huang2022cmua}, TAFIM~\cite{aneja2022tafim}, Dual Defense~\cite{zhang2024dual}, and AdvMark~\cite{wu2024watermarks} proactively inject a watermark-like or adversarial signal at embedding time, so that a later DeepFake generation, face swap, or forensic-detection step is disrupted while some traceability is preserved. Their goal is manipulation prevention by design, not biometric verification under an already watermarked probe.
The closest works that connect watermarking with face recognition \cite{yao2024adversarial} jointly optimize an input perturbation and a watermark message so that a face is correctly recognized before watermarking but fails after the watermark encoder is applied. DIP-Watermark~\cite{zhang2026dip} designs an identity-specific adversarial watermark that degrades unauthorized face recognition while remaining extractable for authorized verification. Both expose the coupling between watermark signals and biometric recognition, but they control the watermarking process itself, by optimizing the message, adapting the encoder, or designing the watermark as an adversarial signal. The setting studied in this paper is inverse and acts at attack time: the adversary receives an already watermarked face image, has no access to the encoder, decoder, or message, and perturbs it so that face recognition fails while the existing watermark remains valid, as in an issuance pipeline, where an authority-watermarked image can be altered downstream to still pass provenance checks while no longer matching its holder, yielding a document that is certified yet biometrically unreliable.

\section{Methodology}
\label{sec:methodology}
 
\subsection{Problem Formulation and Threat Model}
\label{sec:problem_threat}

Let $(\mathbf{x}_t,\mathbf{x}_p)$ denote a mated verification pair, where $\mathbf{x}_t \in [0,1]^{C\times H\times W}$ is the template image and $\mathbf{x}_p \in [0,1]^{C\times H\times W}$ is the probe image. Only the probe is watermarked, since the template is enrolled within the trusted domain of the issuing authority while the probe transits untrusted channels. Given a binary message $\mathbf{m}\in\{0,1\}^{L_m}$, a watermark encoder $\mathcal{W}$ produces:
\begin{equation}
    \mathbf{x}_w = \mathcal{W}(\mathbf{x}_p,\mathbf{m}),
\end{equation}
while a decoder $\mathcal{D}$ estimates the embedded message as $\widehat{\mathbf{m}}=\mathcal{D}(\mathbf{x})$ from a complete image $\mathbf{x}$.

Face verification is performed in a one-to-one setting. Let $\mathcal{A}(\cdot)$ denote the face detection, alignment, and resizing operation used by the recognition system, and let $\mathcal{F}$ be the face-recognition model. The embedding of an image $\mathbf{x}_i$ is:
\begin{equation}
    \mathbf{z}_i = \mathcal{F}(\mathcal{A}(\mathbf{x}_i)) \in \mathbb{R}^{d},
    \qquad \|\mathbf{z}_i\|_2=1,
    \qquad i\in\{t,w,\mathrm{adv}\}.
\end{equation}
The verification score between the template and the watermarked probe is $s_w=\operatorname{sim}(\mathbf{z}_t,\mathbf{z}_w)$, where $\operatorname{sim}(\cdot,\cdot)$ denotes cosine similarity.

The attacker receives the watermarked probe $\mathbf{x}_w$ and generates an adversarial probe:
\begin{equation}
    \mathbf{x}_{\mathrm{adv}} = \operatorname{clip}_{[0,1]}(\mathbf{x}_w + \boldsymbol{\delta}'),
    \qquad
    \boldsymbol{\delta}'=\mathbf{M}\odot\boldsymbol{\delta},
    \qquad
    \|\boldsymbol{\delta}\|_2 \leq \epsilon_2,
\end{equation}
where $\mathbf{M}\in\{0,1\}^{C\times H\times W}$ is the binary face-region mask and $\boldsymbol{\delta}'$ is the effective perturbation applied to the image. The attack minimizes the adversarial verification score:
\begin{equation}
    s_{\mathrm{adv}}
    =
    \operatorname{sim}(\mathbf{z}_t,\mathbf{z}_{\mathrm{adv}})
    \label{eq:sadv}
\end{equation}
below the operational verification threshold, while preserving the hidden message and limiting the distortion introduced with respect to $\mathbf{x}_w$. The objective is therefore not to remove or forge the watermark, but to break the binding between a still-certified image and its biometric identity: the resulting probe passes provenance checks while failing automated verification.
The watermarking system is treated as black-box: the attacker has no access to $\mathcal{W}$, $\mathcal{D}$, $\mathbf{m}$, or gradients associated with watermark recovery. Therefore, bit accuracy cannot be optimized directly. The face-recognition model is white-box, so gradients of the verification loss with respect to the probe image are available. This setting isolates the failure mode studied in this paper: a facial image can remain valid from the perspective of watermark-based traceability while becoming unreliable for biometric verification.

\subsection{Baseline PGD-$L_2$ Evasion Attack}
\label{sec:baseline}

As a baseline, we use an $L_2$ Projected Gradient Descent (PGD) attack~\cite{madry2017towards} optimized exclusively to degrade biometric verification, without explicit mechanisms to preserve the embedded watermark. The adversarial objective is a hinge loss on the cosine similarity between the template and adversarial embeddings:
\begin{equation}
    \mathcal{L}_{\mathrm{adv}}
    =
    \max\left(
    0,
    \operatorname{sim}(\mathbf{z}_t,\mathbf{z}_{\mathrm{adv}})
    -
    \gamma
    \right),
    \label{eq:loss_baseline}
\end{equation}
where $\gamma$ is the target margin. Minimizing Eq.~\eqref{eq:loss_baseline} pushes the adversarial probe away from the template until the similarity falls below the margin. Since no watermark-related term is included, any degradation in message recovery reflects the effect of a purely biometric evasion attack.

At iteration $k$, the perturbation is updated along the normalized gradient direction, restricted to the facial mask, and projected onto the $L_2$ ball:
\begin{equation}
\begin{aligned}
    \mathbf{g}_{\mathrm{base}}^{(k)}
    &=
    \nabla_{\boldsymbol{\delta}}
    \mathcal{L}_{\mathrm{adv}}
    \left(
    \mathbf{z}_t,\mathbf{z}_{\mathrm{adv}}^{(k)}
    \right),
    \\
    \widetilde{\boldsymbol{\delta}}^{(k+1)}
    &=
    \boldsymbol{\delta}^{(k)}
    -
    \alpha_2
    \frac{\mathbf{g}_{\mathrm{base}}^{(k)}}
    {\left\|\mathbf{g}_{\mathrm{base}}^{(k)}\right\|_2+\eta},
    \\
    \boldsymbol{\delta}^{(k+1)}
    &=
    \Pi_{\|\cdot\|_2\leq\epsilon_2}
    \left(
    \mathbf{M}\odot\widetilde{\boldsymbol{\delta}}^{(k+1)}
    \right),
    \\
    \boldsymbol{\delta}'^{(k+1)}
    &=
    \mathbf{M}\odot\boldsymbol{\delta}^{(k+1)},
    \\
    \mathbf{x}_{\mathrm{adv}}^{(k+1)}
    &=
    \operatorname{clip}_{[0,1]}
    \left(
    \mathbf{x}_w+\boldsymbol{\delta}'^{(k+1)}
    \right).
\end{aligned}
\label{eq:pgd_l2_baseline}
\end{equation}
Here, $\alpha_2$ is the step size, $\eta$ is a small numerical constant, and $\Pi_{\|\cdot\|_2\leq\epsilon_2}$ denotes projection onto the $L_2$ ball of radius $\epsilon_2$. Gradients are backpropagated through the same detection, alignment, and resizing pipeline used by the recognition model. The mask $\mathbf{M}$ ensures that the effective perturbation remains confined to the selected facial region. The $L_2$ formulation is adopted because it constrains the total perturbation energy, allowing the attack to concentrate changes on identity-relevant facial regions instead of imposing a uniform per-pixel bound.

\subsection{PGD-$L_2$ with Spatial--Spectral Preservation}
\label{sec:proposed}

\begin{figure*}[t]
\begin{center}
\includegraphics[width=\textwidth]{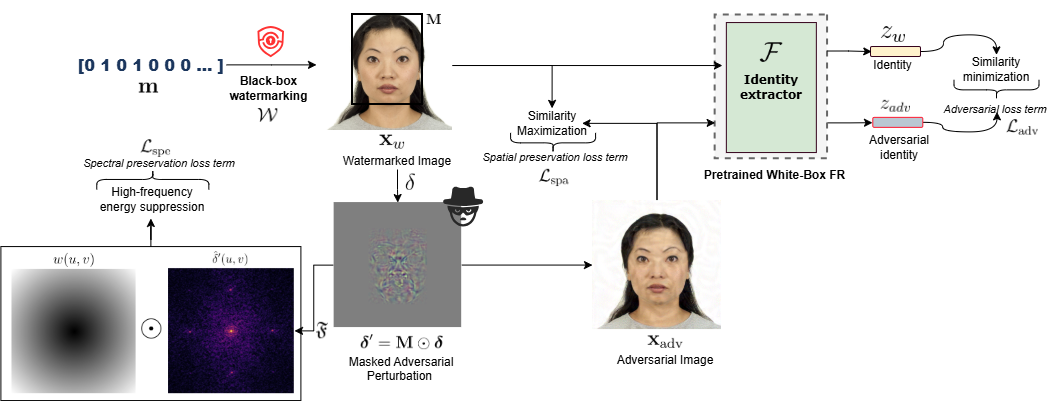}
\end{center}
\caption{Proposed PGD-$L_2$ attack with spatial--spectral preservation. The watermarked probe $\mathbf{x}_w$ is modified by a masked perturbation $\boldsymbol{\delta}'=\mathbf{M}\odot\boldsymbol{\delta}$, producing $\mathbf{x}_{\mathrm{adv}}$. The identity loss reduces the similarity between the template embedding $\mathbf{z}_t$ and the adversarial probe embedding $\mathbf{z}_{\mathrm{adv}}$, while spatial and spectral regularizers constrain the perturbation to preserve image fidelity and reduce interference.}
\label{fig:pipeline}
\end{figure*}

The proposed attack extends the baseline PGD-$L_2$ evasion attack by adding two decoder-free preservation terms to the biometric objective. Figure~\ref{fig:pipeline} illustrates the complete optimization pipeline: the watermarked probe $\mathbf{x}_w$ is modified only through the effective masked perturbation $\boldsymbol{\delta}'$, while the loss combines biometric evasion, spatial fidelity, and spectral preservation. The watermark decoder is never queried during optimization; preservation is encouraged only by constraining the perturbation in the image and frequency domains.

The complete objective is:
\begin{equation}
    \mathcal{L}_{\mathrm{prop}}
    =
    \lambda_{\mathrm{adv}}\mathcal{L}_{\mathrm{adv}}
    +
    \lambda_{\mathrm{spa}}\mathcal{L}_{\mathrm{spa}}
    +
    \lambda_{\mathrm{spe}}\mathcal{L}_{\mathrm{spe}},
    \label{eq:loss_spatial_spectral}
\end{equation}
where $\lambda_{\mathrm{adv}}$, $\lambda_{\mathrm{spa}}$, and $\lambda_{\mathrm{spe}}$ control the relative contribution of biometric evasion, spatial preservation, and spectral preservation. The PGD update follows Eq.~\eqref{eq:pgd_l2_baseline}, replacing the baseline gradient with:
\begin{equation}
    \mathbf{g}_{\mathrm{prop}}^{(k)}
    =
    \nabla_{\boldsymbol{\delta}}
    \mathcal{L}_{\mathrm{prop}}
    \left(
    \mathbf{z}_t,
    \mathbf{z}_{\mathrm{adv}}^{(k)},
    \mathbf{x}_{\mathrm{adv}}^{(k)},
    \mathbf{x}_w
    \right).
    \label{eq:grad_prop}
\end{equation}
Thus, the optimization remains an $L_2$-bounded PGD attack on the watermarked probe, but the update direction is determined by the joint objective rather than by biometric evasion alone.

\paragraph{Spatial preservation.}
The spatial regularizer penalizes pixel-domain deviation between the adversarial probe and the watermarked probe:
\begin{equation}
    \mathcal{L}_{\mathrm{spa}}
    =
    \frac{1}{N}
    \left\|
    \mathbf{x}_{\mathrm{adv}}-\mathbf{x}_w
    \right\|_2^2,
    \qquad N=CHW.
    \label{eq:loss_spatial}
\end{equation}
Although the $L_2$ projection already imposes a maximum perturbation budget, Eq.~\eqref{eq:loss_spatial} biases the optimizer toward using less perturbation energy when the biometric objective can be satisfied with smaller changes. This term promotes visual fidelity and reduces unnecessary pixel-level modifications that may corrupt the embedded message.

\paragraph{Spectral preservation.}
The spectral regularizer constrains where perturbation energy is introduced in the frequency domain. Rather than analyzing the spectrum of the full image, the penalty is applied to the effective perturbation $\boldsymbol{\delta}'$, since the objective is to control the spectral structure of the adversarial change. Let:
\begin{equation}
    \widehat{\boldsymbol{\delta}'}
    =
    \mathfrak{F}(\boldsymbol{\delta}')
    \label{eq:fft_delta}
\end{equation}
denote the two-dimensional discrete Fourier transform of $\boldsymbol{\delta}'$, computed independently for each channel with the zero-frequency component shifted to the center. A normalized radial weight map is defined as:
\begin{equation}
    w(u,v)
    =
    \frac{
    \sqrt{(u-u_c)^2+(v-v_c)^2}
    }
    {r_{\max}+\eta},
    \label{eq:radial_weight}
\end{equation}
where $(u_c,v_c)$ is the spectral center and $r_{\max}$ is the maximum radial distance. The spectral preservation loss is:
\begin{equation}
    \mathcal{L}_{\mathrm{spe}}
    =
    \frac{1}{N}
    \sum_{c,u,v}
    w(u,v)
    \left|
    \widehat{\boldsymbol{\delta}'}(c,u,v)
    \right|^2.
    \label{eq:loss_spectral}
\end{equation}
Unlike an unweighted Fourier-domain energy, which is equivalent to spatial-domain energy up to normalization, Eq.~\eqref{eq:loss_spectral} is frequency-selective. It assigns a larger cost to high-frequency perturbation components, where imperceptible watermark residuals may be sensitive to interference, while remaining independent of the unknown watermark encoder, decoder, and message. A weighted $\ell_1$ variant of Eq.~\eqref{eq:loss_spectral} is considered only in the supplementary ablation; the main attack uses the quadratic spectral penalty unless otherwise stated.

\section{Experimental Protocol}
\label{sec:experimental_setup}

The experimental evaluation is designed to assess whether a watermarked facial image can remain traceable while becoming unreliable for biometric verification. Experiments are conducted on two controlled frontal face datasets representative of identity-document-like acquisition conditions: the Chicago Face Database (CFD)~\cite{ma2015chicago} and FaceLab London~\cite{debruine2017face}. We attack all available probe images in the selected subsets, namely 831 images from CFD and 102 images from FaceLab London. Mated template--probe pairs are built from identities with at least two samples, resulting in 158 identities from CFD and 102 identities from FaceLab London. In each pair, the template and probe share the same identity and differ mainly in facial expression, with changes in smile being the dominant variation. Only the probe image is watermarked and attacked, while the template image remains unchanged.

The biometric verification stage is evaluated using two white-box face recognition models: ArcFace with a ResNet-100 backbone~\cite{deng2019arcface} and FaceNet based on InceptionResnetV1 pretrained on VGGFace2~\cite{schroff2015facenet}. The use of two architectures allows us to evaluate whether the attack behavior remains consistent across different embedding models. Faces and landmarks are detected with MTCNN~\cite{zhang2016joint}, and the detected landmarks are used for face alignment before feature extraction. Aligned faces are resized to $160\times160$ for FaceNet and $112\times112$ for ArcFace. The perturbation $\boldsymbol{\delta}$ is optimized in the image domain. Gradients are propagated through the differentiable preprocessing operations used by the recognition model, while the detected facial region and the corresponding mask $\mathbf{M}$ are kept fixed during each attack.

Before watermark embedding, all probe images are resized to $256\times256$ pixels. Two watermarking architectures are considered to cover different payload regimes. FaceSigns~\cite{neekhara2024facesigns} is used as a low-capacity facial watermarking model embedding 128-bit binary messages, using the authors' released model trained on CelebA. StegaFormer~\cite{yu2024effective} is used as a high-capacity transformer-based watermarking architecture; in our experiments, it is trained at 1 bpp on CelebA-HQ~\cite{karras2017progressive} with random binary messages. During evaluation, a unique binary message is generated for each identity and embedded consistently in the corresponding probe images. 

All attacks are implemented as PGD-$L_2$ optimizations with perturbation budgets $\epsilon_2\in\{0.5,1.0,2.0,3.0\}$. The baseline attack optimizes only the biometric evasion loss $\mathcal{L}_{\mathrm{adv}}$, without spatial or spectral preservation terms, using $K=80$ PGD steps and step size:
\begin{equation}
    \alpha_2 = \frac{2.5\epsilon_2}{K}.
    \label{eq:l2_stepsize_baseline}
\end{equation}
The proposed spatial--spectral attack optimizes the full objective $\mathcal{L}_{\mathrm{prop}}$, which combines biometric evasion with spatial and spectral preservation. Since the preservation terms introduce competing objectives, the proposed attack uses a longer optimization schedule with $K=180$ steps and step size:
\begin{equation}
    \alpha_2 = \frac{15\epsilon_2}{K}.
    \label{eq:l2_stepsize_proposed}
\end{equation}
The preservation weights $\lambda_{\mathrm{spa}}$ and $\lambda_{\mathrm{spe}}$ are varied to analyze the trade-off among biometric evasion, visual fidelity, and watermark recovery, while $\lambda_{\mathrm{adv}}=1$ unless otherwise stated.

For each dataset, watermarking model, face-recognition model, and attack configuration, the evaluation jointly measures biometric evasion, visual fidelity, and watermark recovery. Biometric evasion is quantified through the attack success rate (ASR). Given a verification threshold $\tau$, ASR is defined as the proportion of mated pairs accepted before the attack and rejected after the attack:
\begin{equation}
    \operatorname{ASR}(\tau)
    =
    \frac{1}{|\mathcal{P}|}
    \sum_{i\in\mathcal{P}}
    \mathbb{I}
    \left[
    s_w^{(i)}\geq\tau
    \;\land\;
    s_{\mathrm{adv}}^{(i)}<\tau
    \right],
    \label{eq:asr}
\end{equation}
where $\mathcal{P}$ is the set of mated template--probe pairs, $s_w^{(i)}$ is the similarity between the template and the watermarked probe, and $s_{\mathrm{adv}}^{(i)}$ is the similarity after the attack. Results are reported at two operating points, corresponding to FAR$=1\%$ and FAR$=0.1\%$, denoted as ASR$_1$ and ASR$_{0.1}$, respectively. The corresponding thresholds are estimated separately for each dataset and face-recognition model from non-mated scores before the attack.

Visual fidelity is measured with PSNR. For adversarial perturbations, PSNR is computed between the adversarial watermarked image $\mathbf{x}_{\mathrm{adv}}$ and the original watermarked image $\mathbf{x}_w$, so that the reported value quantifies only the distortion introduced by the attack and not the initial distortion introduced by watermark embedding; pre-attack watermarking quality is instead reported as PSNR between the clean probe $\mathbf{x}_p$ and the watermarked probe $\mathbf{x}_w$. Watermark recovery is measured with the bit accuracy rate, $\operatorname{BAR}(\mathbf{x}) = \frac{1}{L_m}\sum_{j=1}^{L_m}\mathbb{I}[\widehat{m}_j(\mathbf{x})=m_j]$, where $m_j$ is the original message bit and $\widehat{m}_j(\mathbf{x})$ the bit decoded from $\mathbf{x}$; it is computed on both $\mathbf{x}_w$ and $\mathbf{x}_{\mathrm{adv}}$.

\section{Results}
\label{sec:results}

The experimental evaluation is organized around three complementary criteria: biometric evasion, measured by ASR at FAR$=1\%$ and FAR$=0.1\%$; visual fidelity, measured by PSNR; and watermark recovery, measured by BAR. We first report the pre-attack watermarking quality on the selected evaluation images, then analyze the baseline PGD-$L_2$ attack and finally evaluate the proposed spatial--spectral attack. Before applying adversarial perturbations, both watermarking models achieve highly reliable message recovery across the selected datasets, albeit in substantially different visual-fidelity regimes. StegaFormer achieves BAR values between $99.58\%$ and $99.98\%$, with PSNR between $56.97$ and $59.47$ dB. FaceSigns achieves BAR values between $99.91\%$ and $99.99\%$, with PSNR between $31.60$ and $32.62$ dB. These values indicate that both watermarking models provide a stable traceability signal before the attack. The following analyses, therefore, measure the effect of adversarial perturbations on already valid watermarked probes.

\subsection{Baseline PGD-$L_2$ attack}
\label{sec:results_baseline}

\begin{table*}[t]
\begin{center}
\scriptsize
\setlength{\tabcolsep}{3pt}
\renewcommand{\arraystretch}{1.15}
\resizebox{\textwidth}{!}{%
\begin{tabular}{|l|c||c c c c|c c c c||c c c c|c c c c|}
\hline
\multicolumn{2}{|c||}{} 
& \multicolumn{8}{c||}{StegaFormer} 
& \multicolumn{8}{c|}{FaceSigns} \\
\cline{3-18}
\multicolumn{2}{|c||}{} 
& \multicolumn{4}{c|}{ArcFace} 
& \multicolumn{4}{c||}{FaceNet} 
& \multicolumn{4}{c|}{ArcFace} 
& \multicolumn{4}{c|}{FaceNet} \\
\cline{3-6} \cline{7-10} \cline{11-14} \cline{15-18}
Dataset & $\varepsilon_2$ 
& PSNR & BAR & $\text{ASR}_1$ & $\text{ASR}_{0.1}$ 
& PSNR & BAR & $\text{ASR}_1$ & $\text{ASR}_{0.1}$ 
& PSNR & BAR & $\text{ASR}_1$ & $\text{ASR}_{0.1}$ 
& PSNR & BAR & $\text{ASR}_1$ & $\text{ASR}_{0.1}$ \\
\hline\hline

\hline

CFD & $0.5$ 
& $58.96 \pm 0.00$ & $98.20 \pm 0.46$ & $0.0$ & $0.0$ 
& $58.96 \pm 0.00$ & $99.17 \pm 0.45$ & $0.6$ & $2.5$ 
& $58.96 \pm 0.00$ & $99.99 \pm 0.06$ & $0.0$ & $0.0$ 
& $58.96 \pm 0.00$ & $99.99 \pm 0.05$ & $2.5$ & $13.9$ \\

 & $1.0$ 
& $52.94 \pm 0.00$ & $95.23 \pm 0.50$ & $1.9$ & $15.8$ 
& $52.94 \pm 0.00$ & $97.30 \pm 0.29$ & $31.0$ & $69.6$ 
& $52.94 \pm 0.00$ & $99.98 \pm 0.07$ & $10.8$ & $36.7$ 
& $52.94 \pm 0.00$ & $99.98 \pm 0.07$ & $55.7$ & $85.4$ \\

 & $2.0$ 
& $46.94 \pm 0.04$ & $89.49 \pm 0.94$ & $95.6$ & $96.2$ 
& $46.92 \pm 0.00$ & $92.53 \pm 0.74$ & $99.4$ & $100.0$ 
& $46.97 \pm 0.11$ & $99.94 \pm 0.12$ & $99.4$ & $99.4$ 
& $46.92 \pm 0.01$ & $99.87 \pm 0.89$ & $100.0$ & $100.0$ \\

 & $3.0$ 
& $44.34 \pm 0.52$ & $87.01 \pm 0.77$ & $96.2$ & $99.4$ 
& $43.53 \pm 0.17$ & $88.01 \pm 0.86$ & $100.0$ & $100.0$ 
& $44.69 \pm 0.59$ & $99.90 \pm 0.17$ & $99.4$ & $100.0$ 
& $43.72 \pm 0.31$ & $99.59 \pm 1.72$ & $100.0$ & $100.0$ \\

\hline

FaceLab & $0.5$ & $58.96 \pm 0.00$ & $99.76 \pm 0.03$ & $0.0$ & $0.0$ 
& $58.96 \pm 0.00$ & $99.79 \pm 0.02$ & $0.0$ & $2.9$ 
& $58.96 \pm 0.00$ & $99.89 \pm 0.34$ & $0.0$ & $1.0$ 
& $58.96 \pm 0.00$ & $99.91 \pm 0.27$ & $3.9$ & $13.7$ \\

London & $1.0$ 
& $52.94 \pm 0.00$ & $98.44 \pm 0.14$ & $34.3$ & $64.7$ 
& $52.94 \pm 0.00$ & $98.68 \pm 0.10$ & $50.0$ & $84.3$ 
& $52.94 \pm 0.00$ & $99.87 \pm 0.41$ & $62.7$ & $83.3$ 
& $52.94 \pm 0.00$ & $99.93 \pm 0.23$ & $82.4$ & $98.0$ \\

 & $2.0$ 
& $47.94 \pm 0.45$ & $94.18 \pm 0.27$ & $100.0$ & $100.0$ 
& $47.21 \pm 0.29$ & $94.53 \pm 0.40$ & $100.0$ & $100.0$ 
& $48.22 \pm 0.44$ & $99.80 \pm 0.71$ & $100.0$ & $100.0$ 
& $47.71 \pm 0.45$ & $99.90 \pm 0.28$ & $100.0$ & $100.0$ \\

 & $3.0$ 
& $46.83 \pm 0.50$ & $93.55 \pm 0.34$ & $100.0$ & $100.0$ 
& $45.69 \pm 0.58$ & $93.56 \pm 0.34$ & $100.0$ & $100.0$ 
& $47.15 \pm 0.44$ & $99.76 \pm 0.90$ & $100.0$ & $100.0$ 
& $46.57 \pm 0.55$ & $99.89 \pm 0.32$ & $100.0$ & $100.0$\\
\hline
\end{tabular}%
}
\end{center}
\caption{Baseline white-box PGD-$L_2$ attack results on \textbf{CFD} and \textbf{FaceLab London}. $K{=}80$ steps, $\lambda_{\mathrm{adv}}{=}1$, $\lambda_{\mathrm{spa}}{=}0$ and $\lambda_{\mathrm{spe}}{=}0$. PSNR in dB, BAR and ASR in \%. Mean $\pm$ std reported for PSNR and BAR.}
\label{tab:baseline}
\end{table*}

Table~\ref{tab:baseline} reports the results of the baseline PGD-$L_2$ attack on CFD and FaceLab London. The baseline optimizes only the adversarial verification loss and therefore provides the reference condition for assessing whether watermark recovery can survive a purely biometric evasion objective. Across all datasets, recognition models, and watermarking schemes, the attack is highly effective against face verification: at $\varepsilon_2=2.0$, ASR$_{0.1}$ exceeds $96\%$ in all configurations. This confirms that watermark embedding alone does not prevent adversarial evasion of the downstream face-recognition system.
The impact on watermark recovery depends strongly on the watermarking architecture. For FaceSigns, BAR remains above $99.7\%$ in all baseline configurations, even at the largest perturbation budgets. A standard adversarial perturbation restricted to the facial region is therefore sufficient to produce a probe that is rejected by the biometric system while remaining almost perfectly traceable. This result exposes the watermark--identity gap already under a conventional biometric attack, particularly for low-capacity facial watermarking.

StegaFormer is more sensitive to the same perturbations. On CFD, BAR decreases to $89.49$--$87.01\%$ for $\varepsilon_2\geq2.0$ when attacking ArcFace, while on FaceLab London it remains above $93\%$ for both recognition models. This behavior is consistent with the higher embedding density of the 1 bpp configuration, where more information is distributed across the image and message recovery becomes more sensitive to pixel-level adversarial changes. Nevertheless, the watermark is not destroyed: BAR remains far above random chance even in the most degraded cases. These baseline results motivate the preservation terms introduced in Sec.~\ref{sec:proposed}, whose purpose is not to make evasion possible, but to improve watermark recovery while maintaining high attack success.

\subsection{Proposed attack with spatial--spectral terms}
```latex
\begin{table*}[t]
\centering
\tiny
\setlength{\tabcolsep}{3pt}
\renewcommand{\arraystretch}{1.15}

\resizebox{\textwidth}{!}{%
\begin{tabular}{|cccc|cccc|cccc|}
\hline
\multicolumn{4}{|c|}{} &
\multicolumn{4}{|c|}{$\lambda_{\mathrm{spa}}=1,\;\lambda_{\mathrm{spe}}=0$} &
\multicolumn{4}{c|}{$\lambda_{\mathrm{spa}}=1,\;\lambda_{\mathrm{spe}}=1$} \\
\hline
\multicolumn{1}{|c|}{Watermarking model}
& \multicolumn{1}{c|}{FR model}
& \multicolumn{1}{c|}{$\varepsilon_2$}
& Dataset
& \multicolumn{1}{c|}{PSNR}
& \multicolumn{1}{c|}{BAR}
& \multicolumn{1}{c|}{$\mathrm{ASR}_1$}
& $\mathrm{ASR}_{0.1}$
& \multicolumn{1}{c|}{PSNR}
& \multicolumn{1}{c|}{BAR}
& \multicolumn{1}{c|}{$\mathrm{ASR}_1$}
& $\mathrm{ASR}_{0.1}$ \\
\hline

\multicolumn{1}{|c|}{\multirow{16}{*}{FaceSigns}}
& \multicolumn{1}{c|}{\multirow{8}{*}{ArcFace}}
& \multicolumn{1}{c|}{\multirow{2}{*}{0.5}}
& CFD
& \multicolumn{1}{c|}{58.96 $\pm$ 0.00}
& \multicolumn{1}{c|}{99.99 $\pm$ 0.06}
& \multicolumn{1}{c|}{0.0}
& 0.0
& \multicolumn{1}{c|}{58.96 $\pm$ 0.00}
& \multicolumn{1}{c|}{99.99 $\pm$ 0.06}
& \multicolumn{1}{c|}{0.0}
& 0.0 \\ \cline{4-12}

\multicolumn{1}{|c|}{}
& \multicolumn{1}{c|}{}
& \multicolumn{1}{c|}{}
& FaceLab London
& \multicolumn{1}{c|}{58.96 $\pm$ 0.00}
& \multicolumn{1}{c|}{99.92 $\pm$ 0.27}
& \multicolumn{1}{c|}{0.0}
& 1.0
& \multicolumn{1}{c|}{58.96 $\pm$ 0.00}
& \multicolumn{1}{c|}{99.92 $\pm$ 0.27}
& \multicolumn{1}{c|}{0.0}
& 1.0 \\ \cline{3-12}

\multicolumn{1}{|c|}{}
& \multicolumn{1}{c|}{}
& \multicolumn{1}{c|}{\multirow{2}{*}{1.0}}
& CFD
& \multicolumn{1}{c|}{52.94 $\pm$ 0.01}
& \multicolumn{1}{c|}{99.98 $\pm$ 0.06}
& \multicolumn{1}{c|}{37.3}
& 69.6
& \multicolumn{1}{c|}{52.94 $\pm$ 0.01}
& \multicolumn{1}{c|}{99.98 $\pm$ 0.06}
& \multicolumn{1}{c|}{38.0}
& 69.6 \\ \cline{4-12}

\multicolumn{1}{|c|}{}
& \multicolumn{1}{c|}{}
& \multicolumn{1}{c|}{}
& FaceLab London
& \multicolumn{1}{c|}{53.09 $\pm$ 0.20}
& \multicolumn{1}{c|}{99.89 $\pm$ 0.34}
& \multicolumn{1}{c|}{90.2}
& 95.1
& \multicolumn{1}{c|}{53.09 $\pm$ 0.18}
& \multicolumn{1}{c|}{99.89 $\pm$ 0.34}
& \multicolumn{1}{c|}{90.2}
& 95.1 \\ \cline{3-12}

\multicolumn{1}{|c|}{}
& \multicolumn{1}{c|}{}
& \multicolumn{1}{c|}{\multirow{2}{*}{2.0}}
& CFD
& \multicolumn{1}{c|}{49.35 $\pm$ 1.00}
& \multicolumn{1}{c|}{99.95 $\pm$ 0.11}
& \multicolumn{1}{c|}{97.5}
& 98.7
& \multicolumn{1}{c|}{49.33 $\pm$ 1.03}
& \multicolumn{1}{c|}{99.96 $\pm$ 0.11}
& \multicolumn{1}{c|}{99.4}
& 99.4 \\ \cline{4-12}

\multicolumn{1}{|c|}{}
& \multicolumn{1}{c|}{}
& \multicolumn{1}{c|}{}
& FaceLab London
& \multicolumn{1}{c|}{51.12 $\pm$ 1.24}
& \multicolumn{1}{c|}{99.86 $\pm$ 0.56}
& \multicolumn{1}{c|}{100.0}
& 100.0
& \multicolumn{1}{c|}{51.28 $\pm$ 1.05}
& \multicolumn{1}{c|}{99.85 $\pm$ 0.56}
& \multicolumn{1}{c|}{100.0}
& 100.0 \\ \cline{3-12}

\multicolumn{1}{|c|}{}
& \multicolumn{1}{c|}{}
& \multicolumn{1}{c|}{\multirow{2}{*}{3.0}}
& CFD
& \multicolumn{1}{c|}{49.08 $\pm$ 1.35}
& \multicolumn{1}{c|}{99.95 $\pm$ 0.11}
& \multicolumn{1}{c|}{98.1}
& 99.4
& \multicolumn{1}{c|}{49.05 $\pm$ 1.41}
& \multicolumn{1}{c|}{99.95 $\pm$ 0.11}
& \multicolumn{1}{c|}{99.4}
& 99.4 \\ \cline{4-12}

\multicolumn{1}{|c|}{}
& \multicolumn{1}{c|}{}
& \multicolumn{1}{c|}{}
& FaceLab London
& \multicolumn{1}{c|}{51.42 $\pm$ 0.81}
& \multicolumn{1}{c|}{99.89 $\pm$ 0.34}
& \multicolumn{1}{c|}{100.0}
& 100.0
& \multicolumn{1}{c|}{51.00 $\pm$ 1.35}
& \multicolumn{1}{c|}{99.79 $\pm$ 0.78}
& \multicolumn{1}{c|}{100.0}
& 100.0 \\ \cline{2-12}

\multicolumn{1}{|c|}{}
& \multicolumn{1}{c|}{\multirow{8}{*}{FaceNet}}
& \multicolumn{1}{c|}{\multirow{2}{*}{0.5}}
& CFD
& \multicolumn{1}{c|}{58.96 $\pm$ 0.00}
& \multicolumn{1}{c|}{99.99 $\pm$ 0.05}
& \multicolumn{1}{c|}{3.8}
& 19.6
& \multicolumn{1}{c|}{58.96 $\pm$ 0.00}
& \multicolumn{1}{c|}{99.99 $\pm$ 0.05}
& \multicolumn{1}{c|}{3.8}
& 19.6 \\ \cline{4-12}

\multicolumn{1}{|c|}{}
& \multicolumn{1}{c|}{}
& \multicolumn{1}{c|}{}
& FaceLab London
& \multicolumn{1}{c|}{58.96 $\pm$ 0.00}
& \multicolumn{1}{c|}{99.92 $\pm$ 0.23}
& \multicolumn{1}{c|}{4.9}
& 24.5
& \multicolumn{1}{c|}{58.96 $\pm$ 0.00}
& \multicolumn{1}{c|}{99.92 $\pm$ 0.23}
& \multicolumn{1}{c|}{4.9}
& 25.5 \\ \cline{3-12}

\multicolumn{1}{|c|}{}
& \multicolumn{1}{c|}{}
& \multicolumn{1}{c|}{\multirow{2}{*}{1.0}}
& CFD
& \multicolumn{1}{c|}{52.94 $\pm$ 0.00}
& \multicolumn{1}{c|}{99.98 $\pm$ 0.07}
& \multicolumn{1}{c|}{70.3}
& 90.5
& \multicolumn{1}{c|}{52.94 $\pm$ 0.00}
& \multicolumn{1}{c|}{99.98 $\pm$ 0.06}
& \multicolumn{1}{c|}{70.9}
& 90.5 \\ \cline{4-12}

\multicolumn{1}{|c|}{}
& \multicolumn{1}{c|}{}
& \multicolumn{1}{c|}{}
& FaceLab London
& \multicolumn{1}{c|}{53.01 $\pm$ 0.13}
& \multicolumn{1}{c|}{99.88 $\pm$ 0.41}
& \multicolumn{1}{c|}{97.1}
& 100.0
& \multicolumn{1}{c|}{53.01 $\pm$ 0.13}
& \multicolumn{1}{c|}{99.87 $\pm$ 0.41}
& \multicolumn{1}{c|}{97.1}
& 100.0 \\ \cline{3-12}

\multicolumn{1}{|c|}{}
& \multicolumn{1}{c|}{}
& \multicolumn{1}{c|}{\multirow{2}{*}{2.0}}
& CFD
& \multicolumn{1}{c|}{47.78 $\pm$ 0.55}
& \multicolumn{1}{c|}{99.92 $\pm$ 0.56}
& \multicolumn{1}{c|}{100.0}
& 100.0
& \multicolumn{1}{c|}{47.75 $\pm$ 0.57}
& \multicolumn{1}{c|}{99.84 $\pm$ 0.97}
& \multicolumn{1}{c|}{100.0}
& 100.0 \\ \cline{4-12}

\multicolumn{1}{|c|}{}
& \multicolumn{1}{c|}{}
& \multicolumn{1}{c|}{}
& FaceLab London
& \multicolumn{1}{c|}{50.03 $\pm$ 1.17}
& \multicolumn{1}{c|}{99.86 $\pm$ 0.49}
& \multicolumn{1}{c|}{100.0}
& 100.0
& \multicolumn{1}{c|}{50.20 $\pm$ 1.15}
& \multicolumn{1}{c|}{99.79 $\pm$ 0.82}
& \multicolumn{1}{c|}{100.0}
& 100.0 \\ \cline{3-12}

\multicolumn{1}{|c|}{}
& \multicolumn{1}{c|}{}
& \multicolumn{1}{c|}{\multirow{2}{*}{3.0}}
& CFD
& \multicolumn{1}{c|}{46.81 $\pm$ 1.25}
& \multicolumn{1}{c|}{99.92 $\pm$ 0.21}
& \multicolumn{1}{c|}{100.0}
& 100.0
& \multicolumn{1}{c|}{46.91 $\pm$ 1.20}
& \multicolumn{1}{c|}{99.74 $\pm$ 1.42}
& \multicolumn{1}{c|}{100.0}
& 100.0 \\ \cline{4-12}

\multicolumn{1}{|c|}{}
& \multicolumn{1}{c|}{}
& \multicolumn{1}{c|}{}
& FaceLab London
& \multicolumn{1}{c|}{50.11 $\pm$ 1.39}
& \multicolumn{1}{c|}{99.83 $\pm$ 0.71}
& \multicolumn{1}{c|}{100.0}
& 100.0
& \multicolumn{1}{c|}{50.44 $\pm$ 1.36}
& \multicolumn{1}{c|}{99.78 $\pm$ 0.86}
& \multicolumn{1}{c|}{100.0}
& 100.0 \\ \hline

\multicolumn{1}{|c|}{\multirow{16}{*}{StegaFormer}}
& \multicolumn{1}{c|}{\multirow{8}{*}{ArcFace}}
& \multicolumn{1}{c|}{\multirow{2}{*}{0.5}}
& CFD
& \multicolumn{1}{c|}{58.96 $\pm$ 0.00}
& \multicolumn{1}{c|}{98.43 $\pm$ 0.47}
& \multicolumn{1}{c|}{0.0}
& 0.0
& \multicolumn{1}{c|}{58.96 $\pm$ 0.00}
& \multicolumn{1}{c|}{98.44 $\pm$ 0.47}
& \multicolumn{1}{c|}{0.0}
& 0.0 \\ \cline{4-12}

\multicolumn{1}{|c|}{}
& \multicolumn{1}{c|}{}
& \multicolumn{1}{c|}{}
& FaceLab London
& \multicolumn{1}{c|}{58.96 $\pm$ 0.00}
& \multicolumn{1}{c|}{99.85 $\pm$ 0.02}
& \multicolumn{1}{c|}{0.0}
& 0.0
& \multicolumn{1}{c|}{58.96 $\pm$ 0.00}
& \multicolumn{1}{c|}{99.85 $\pm$ 0.02}
& \multicolumn{1}{c|}{0.0}
& 0.0 \\ \cline{3-12}

\multicolumn{1}{|c|}{}
& \multicolumn{1}{c|}{}
& \multicolumn{1}{c|}{\multirow{2}{*}{1.0}}
& CFD
& \multicolumn{1}{c|}{52.94 $\pm$ 0.00}
& \multicolumn{1}{c|}{96.13 $\pm$ 0.54}
& \multicolumn{1}{c|}{14.6}
& 44.3
& \multicolumn{1}{c|}{52.94 $\pm$ 0.00}
& \multicolumn{1}{c|}{96.13 $\pm$ 0.54}
& \multicolumn{1}{c|}{14.6}
& 44.3 \\ \cline{4-12}

\multicolumn{1}{|c|}{}
& \multicolumn{1}{c|}{}
& \multicolumn{1}{c|}{}
& FaceLab London
& \multicolumn{1}{c|}{53.00 $\pm$ 0.13}
& \multicolumn{1}{c|}{99.29 $\pm$ 0.09}
& \multicolumn{1}{c|}{72.5}
& 83.3
& \multicolumn{1}{c|}{53.00 $\pm$ 0.12}
& \multicolumn{1}{c|}{99.32 $\pm$ 0.11}
& \multicolumn{1}{c|}{72.5}
& 82.4 \\ \cline{3-12}

\multicolumn{1}{|c|}{}
& \multicolumn{1}{c|}{}
& \multicolumn{1}{c|}{\multirow{2}{*}{2.0}}
& CFD
& \multicolumn{1}{c|}{48.74 $\pm$ 0.84}
& \multicolumn{1}{c|}{93.62 $\pm$ 1.13}
& \multicolumn{1}{c|}{94.9}
& 96.2
& \multicolumn{1}{c|}{48.61 $\pm$ 0.86}
& \multicolumn{1}{c|}{94.53 $\pm$ 1.57}
& \multicolumn{1}{c|}{95.6}
& 96.2 \\ \cline{4-12}

\multicolumn{1}{|c|}{}
& \multicolumn{1}{c|}{}
& \multicolumn{1}{c|}{}
& FaceLab London
& \multicolumn{1}{c|}{50.92 $\pm$ 1.11}
& \multicolumn{1}{c|}{98.43 $\pm$ 1.44}
& \multicolumn{1}{c|}{100.0}
& 100.0
& \multicolumn{1}{c|}{50.81 $\pm$ 1.05}
& \multicolumn{1}{c|}{98.70 $\pm$ 1.46}
& \multicolumn{1}{c|}{100.0}
& 100.0 \\ \cline{3-12}

\multicolumn{1}{|c|}{}
& \multicolumn{1}{c|}{}
& \multicolumn{1}{c|}{\multirow{2}{*}{3.0}}
& CFD
& \multicolumn{1}{c|}{48.26 $\pm$ 1.39}
& \multicolumn{1}{c|}{92.91 $\pm$ 2.19}
& \multicolumn{1}{c|}{94.9}
& 95.6
& \multicolumn{1}{c|}{48.19 $\pm$ 1.44}
& \multicolumn{1}{c|}{93.99 $\pm$ 2.44}
& \multicolumn{1}{c|}{96.2}
& 96.8 \\ \cline{4-12}

\multicolumn{1}{|c|}{}
& \multicolumn{1}{c|}{}
& \multicolumn{1}{c|}{}
& FaceLab London
& \multicolumn{1}{c|}{50.55 $\pm$ 1.42}
& \multicolumn{1}{c|}{98.28 $\pm$ 1.76}
& \multicolumn{1}{c|}{100.0}
& 100.0
& \multicolumn{1}{c|}{50.60 $\pm$ 1.17}
& \multicolumn{1}{c|}{98.71 $\pm$ 1.64}
& \multicolumn{1}{c|}{100.0}
& 100.0 \\ \cline{2-12}

\multicolumn{1}{|c|}{}
& \multicolumn{1}{c|}{\multirow{8}{*}{FaceNet}}
& \multicolumn{1}{c|}{\multirow{2}{*}{0.5}}
& CFD
& \multicolumn{1}{c|}{58.96 $\pm$ 0.00}
& \multicolumn{1}{c|}{99.40 $\pm$ 0.45}
& \multicolumn{1}{c|}{1.9}
& 5.7
& \multicolumn{1}{c|}{58.96 $\pm$ 0.00}
& \multicolumn{1}{c|}{99.41 $\pm$ 0.44}
& \multicolumn{1}{c|}{1.9}
& 5.7 \\ \cline{4-12}

\multicolumn{1}{|c|}{}
& \multicolumn{1}{c|}{}
& \multicolumn{1}{c|}{}
& FaceLab London
& \multicolumn{1}{c|}{58.96 $\pm$ 0.00}
& \multicolumn{1}{c|}{99.89 $\pm$ 0.01}
& \multicolumn{1}{c|}{2.0}
& 4.9
& \multicolumn{1}{c|}{58.96 $\pm$ 0.00}
& \multicolumn{1}{c|}{99.89 $\pm$ 0.01}
& \multicolumn{1}{c|}{2.0}
& 4.9 \\ \cline{3-12}

\multicolumn{1}{|c|}{}
& \multicolumn{1}{c|}{}
& \multicolumn{1}{c|}{\multirow{2}{*}{1.0}}
& CFD
& \multicolumn{1}{c|}{52.94 $\pm$ 0.00}
& \multicolumn{1}{c|}{98.80 $\pm$ 0.41}
& \multicolumn{1}{c|}{54.4}
& 81.0
& \multicolumn{1}{c|}{52.94 $\pm$ 0.00}
& \multicolumn{1}{c|}{98.80 $\pm$ 0.41}
& \multicolumn{1}{c|}{53.8}
& 81.0 \\ \cline{4-12}

\multicolumn{1}{|c|}{}
& \multicolumn{1}{c|}{}
& \multicolumn{1}{c|}{}
& FaceLab London
& \multicolumn{1}{c|}{52.95 $\pm$ 0.04}
& \multicolumn{1}{c|}{99.49 $\pm$ 0.05}
& \multicolumn{1}{c|}{79.4}
& 98.0
& \multicolumn{1}{c|}{52.95 $\pm$ 0.03}
& \multicolumn{1}{c|}{99.50 $\pm$ 0.05}
& \multicolumn{1}{c|}{79.4}
& 98.0 \\ \cline{3-12}

\multicolumn{1}{|c|}{}
& \multicolumn{1}{c|}{}
& \multicolumn{1}{c|}{\multirow{2}{*}{2.0}}
& CFD
& \multicolumn{1}{c|}{47.30 $\pm$ 0.38}
& \multicolumn{1}{c|}{97.07 $\pm$ 0.52}
& \multicolumn{1}{c|}{100.0}
& 100.0
& \multicolumn{1}{c|}{47.29 $\pm$ 0.37}
& \multicolumn{1}{c|}{97.55 $\pm$ 0.64}
& \multicolumn{1}{c|}{100.0}
& 100.0 \\ \cline{4-12}

\multicolumn{1}{|c|}{}
& \multicolumn{1}{c|}{}
& \multicolumn{1}{c|}{}
& FaceLab London
& \multicolumn{1}{c|}{49.71 $\pm$ 0.68}
& \multicolumn{1}{c|}{98.56 $\pm$ 1.12}
& \multicolumn{1}{c|}{100.0}
& 100.0
& \multicolumn{1}{c|}{49.50 $\pm$ 0.94}
& \multicolumn{1}{c|}{98.75 $\pm$ 1.38}
& \multicolumn{1}{c|}{100.0}
& 100.0 \\ \cline{3-12}

\multicolumn{1}{|c|}{}
& \multicolumn{1}{c|}{}
& \multicolumn{1}{c|}{\multirow{2}{*}{3.0}}
& CFD
& \multicolumn{1}{c|}{46.01 $\pm$ 0.99}
& \multicolumn{1}{c|}{95.38 $\pm$ 2.49}
& \multicolumn{1}{c|}{100.0}
& 100.0
& \multicolumn{1}{c|}{45.98 $\pm$ 1.05}
& \multicolumn{1}{c|}{96.39 $\pm$ 2.76}
& \multicolumn{1}{c|}{100.0}
& 100.0 \\ \cline{4-12}

\multicolumn{1}{|c|}{}
& \multicolumn{1}{c|}{}
& \multicolumn{1}{c|}{}
& FaceLab London
& \multicolumn{1}{c|}{48.86 $\pm$ 1.57}
& \multicolumn{1}{c|}{97.46 $\pm$ 2.12}
& \multicolumn{1}{c|}{100.0}
& 100.0
& \multicolumn{1}{c|}{48.78 $\pm$ 1.51}
& \multicolumn{1}{c|}{97.74 $\pm$ 2.38}
& \multicolumn{1}{c|}{100.0}
& 100.0 \\ \hline

\end{tabular}%
}

\caption{Proposed method results for \textbf{FaceSigns} and
\textbf{StegaFormer} under ArcFace and FaceNet white-box FR models.
PGD-$L_2$ attack, $K=180$ steps, $\lambda_{\mathrm{adv}}=1$.
PSNR in dB, BAR and ASR in \% (mean $\pm$ std reported).}
\label{tab:method}
\end{table*}
```

\begin{figure}[t]
\begin{center}
\includegraphics[width=\linewidth]{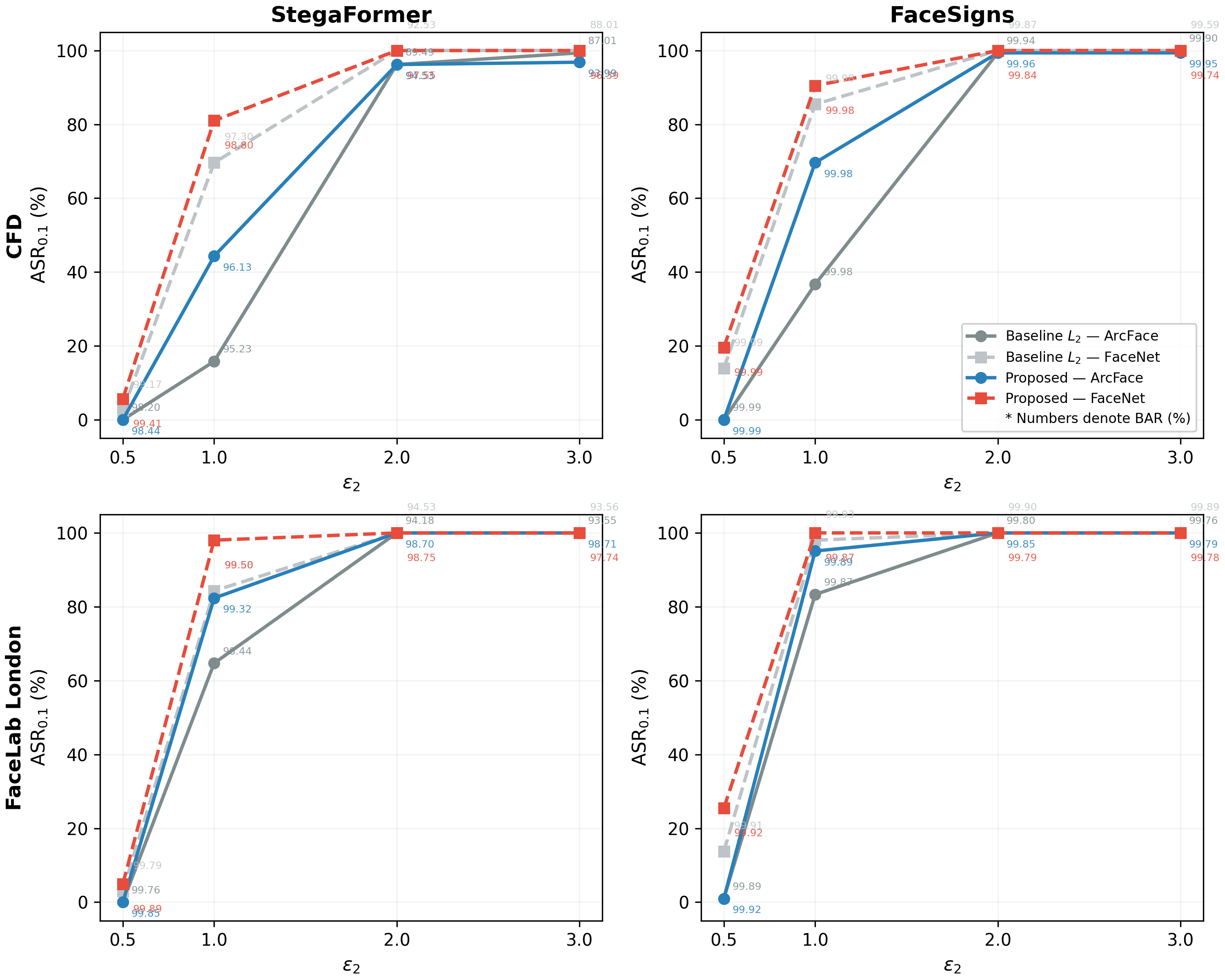}
\end{center}
\caption{$\mathrm{ASR}_{0.1}$ vs. $\epsilon_2$ for the baseline PGD-L2 (gray) and the proposed spatial--spectral attack (blue and red); solid lines denote ArcFace, dashed FaceNet. Numbers next to each point report the corresponding BAR (\%).}
\label{fig:asr_vs_eps}
\end{figure}

Table~\ref{tab:method} reports the proposed PGD-$L_2$ attack with spatial-only preservation ($\lambda_{\mathrm{spa}}=1$, $\lambda_{\mathrm{spe}}=0$) and spatial--spectral preservation ($\lambda_{\mathrm{spa}}=1$, $\lambda_{\mathrm{spe}}=1$). All configurations use $\lambda_{\mathrm{adv}}=1$. For FaceSigns, the attack still preserves the watermark almost perfectly across all settings. BAR remains above $99.7\%$ for all configurations, while ASR$_{0.1}$ reaches $100\%$ for $\varepsilon_2 \geq 2.0$. However, incorporating the spectral term at higher budgets ($\varepsilon_2 \geq 2.0$) causes a slight degradation in BAR. Because FaceSigns introduces visible artifacts in low and mid-frequencies, the spectral constraint allows the adversarial perturbation to alter these specific frequency bands, marginally impacting watermark extraction (see visual perturbations in Figure \ref{fig:visual_examples}). 
StegaFormer represents a more challenging setting. Compared with the baseline, the proposed preservation terms substantially improve BAR while maintaining high evasion rates. On CFD at $\varepsilon_2=2.0$, BAR increases from $89.49\%$ to $93.62\%$ with spatial preservation and to $94.53\%$ with spatial--spectral preservation when attacking ArcFace. For FaceNet, the corresponding increase is from $92.53\%$ to $97.07\%$ and $97.55\%$. A similar trend is observed on FaceLab London, where BAR reaches $98.70$--$98.75\%$ at the same perturbation budget. These results show that the proposed regularization is most useful for high-capacity and high imperceptibility watermarking, where the embedded message is more sensitive to adversarial pixel-level changes. The spectral term provides a consistent, yet moderate, additional gain, especially at larger perturbation budgets. At $\varepsilon_2=3.0$ on CFD, spatial--spectral preservation improves BAR from $92.91\%$ to $93.99\%$ for ArcFace and from $95.38\%$ to $96.39\%$ for FaceNet. This suggests that most of the preservation effect is already captured by the spatial term, while the spectral term further regularizes the perturbation when the attack has more energy to distribute. Importantly, the improved watermark recovery does not substantially reduce the effectiveness of attacks. At $\varepsilon_2=2.0$, the proposed attack reaches ASR$_{0.1}\geq96.2\%$ on CFD for ArcFace and $100\%$ for FaceNet, while all FaceLab London configurations reach $100\%$ ASR$_{0.1}$. Thus, the proposed objective simultaneously satisfies the three target conditions: high biometric evasion, high watermark recovery, and high visual fidelity. Figure~\ref{fig:asr_vs_eps} summarizes this trade-off by showing that the proposed attack improve ASR and BAR with respect to the baselines.

\subsection{The watermark--identity gap}

\begin{figure}
\begin{center}
\includegraphics[width=\columnwidth]{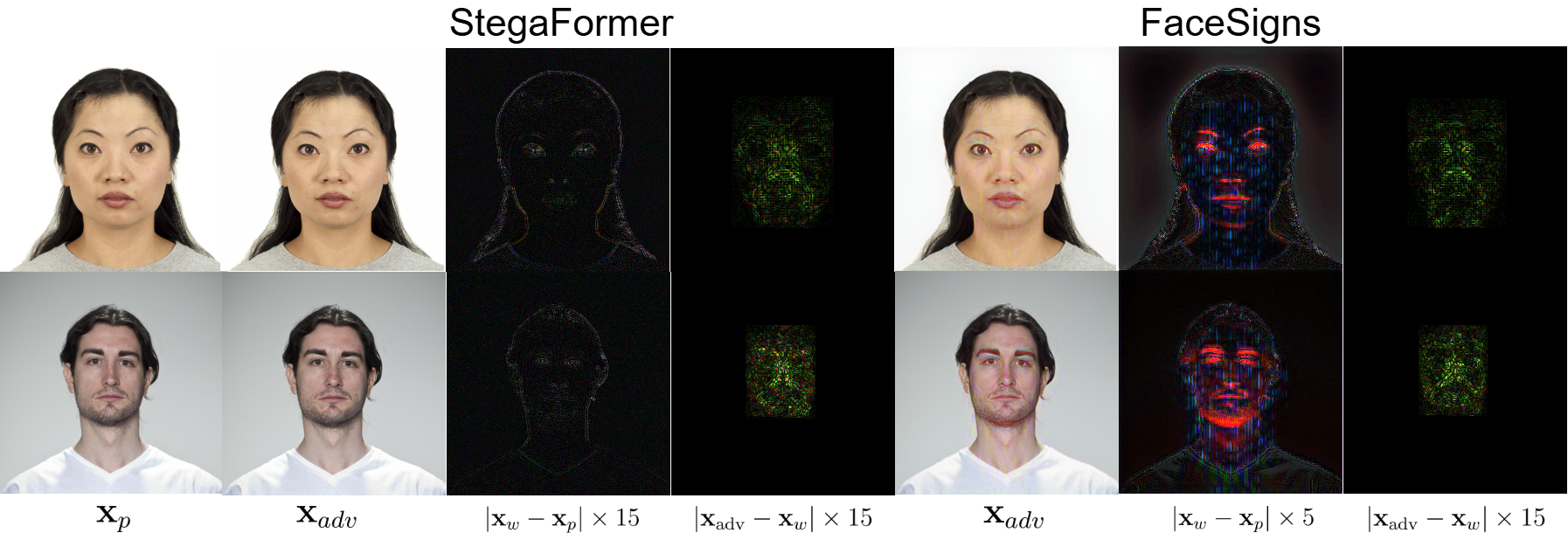}
\end{center}
\caption{Watermark-preserving attack on one identity from CFD (top) and
one from FaceLab London (bottom), at $\epsilon_2 = 2.0$ against ArcFace.
The probe $x_p$ is shared by both watermarking models; for each model we
show $x_{\mathrm{adv}}$, the watermark residual, and the adversarial
perturbation. Residuals are amplified $\times 15$ for StegaFormer and
$\times 5$ for FaceSigns to reach comparable visibility.}
\label{fig:visual_examples}
\end{figure}

The combined results reveal a consistent watermark--identity gap. At perturbation budgets of $\varepsilon_2 \geq 2.0$, the proposed attack achieves evasion rates above $92\%$ at FAR$=0.1\%$, while preserving BAR above $93\%$ for StegaFormer and above $99\%$ for FaceSigns. At $\varepsilon_2=2.0$, visual quality also remains above $47$~dB PSNR across both datasets and recognition models. Thus, the attacked probes simultaneously satisfy three conditions: they are biometrically rejected, their hidden watermark remains recoverable, and their visual distortion remains limited. Figure~\ref{fig:visual_examples} illustrates this behavior, showing adversarial probes that remain visually close to the watermarked images while the perturbation is confined to the facial region.

The practical implication is that watermark validity alone is insufficient to certify biometric reliability. The recovered BAR after the proposed attack lies within, and in several cases above, the recovery levels that FaceSigns was designed to tolerate under benign image operations such as compression and social-media filtering~\cite{neekhara2024facesigns}. In particular, FaceSigns remains above $99.7\%$ BAR under all tested adversarial configurations, which is comparable to or higher than the recovery reported for standard benign transformations in its original evaluation. Similarly, StegaFormer can retain near-unperturbed recovery on FaceLab London at $\varepsilon_2=1.0$, reaching $99.32\%$ and $99.50\%$ for ArcFace/Facenet, respectively, while already achieving high biometric evasion ($82.4\%$ and $98\%$). A verifier relying only on successful message recovery would therefore have no direct evidence that the biometric content has been adversarially degraded.

The two watermarking schemes expose complementary aspects of the vulnerability. For FaceSigns, the gap already appears under the baseline PGD attack. The low-capacity watermark remains almost perfectly recoverable even without explicit preservation terms. For StegaFormer, the higher payload density makes recovery more sensitive to adversarial perturbations, and the proposed spatial--spectral regularization improves BAR by several percentage points while maintaining comparable evasion rates. In both cases, the central conclusion is the same: provenance and identity are independent properties of a facial image, and a valid watermark should not be treated as sufficient evidence that the image remains reliable for face recognition.

\section{Conclusions}
\label{sec:conclusions}

This work introduced a watermark-preserving face evasion attack that exposes a structural gap in watermark-based facial certification. The proposed PGD-$L_2$ attack uses spatial and spectral preservation terms to generate adversarial probes that remain traceable through their hidden watermark, while becoming unreliable for face verification. Across two controlled face datasets, two watermarking architectures, and two face recognition models, the attack achieves high evasion rates at operational thresholds while preserving high watermark recovery and visual quality. The results show that this gap depends on the watermarking regime. For FaceSigns, the watermark remains almost perfectly recoverable even under a standard PGD attack, showing that the vulnerability can already emerge without explicit preservation terms. For StegaFormer, the higher payload density makes recovery more sensitive to adversarial perturbations, but the proposed spatial--spectral regularization substantially improves BAR while maintaining high attack success. In both cases, successful watermark recovery does not imply that the image remains biometrically reliable. The evaluated watermarking models were not designed to resist this threat; the results expose a limitation of certification pipelines that treat watermark validity as a proxy for biometric trustworthiness. The current attack assumes white-box access to the face recognition model, and transferability to unseen models remains an important direction for future work. From a defensive perspective, these findings motivate certification pipelines that treat watermark validity and biometric reliability as complementary security properties rather than interchangeable indicators of trustworthiness. Developing and evaluating integrated defenses capable of jointly assessing provenance preservation and biometric consistency under adversarial conditions remains an important direction for future research. The adversarial perturbation is designed to remain below semi-fragile detection thresholds, highlighting the need to further investigate whether defenses that rely solely on watermark fragility are sufficient against this class of attacks. More broadly, this work provides evidence that provenance and identity are independent properties of a facial image, and that systems designed to guarantee one do not automatically guarantee the other.
\section*{Acknowledgment}

Davide Ghiani's PhD grant is partly funded by Dedem SpA under the PNRR program. The authors thank Federico Lama for his valuable advice.
{\small
\bibliographystyle{ieee}
\bibliography{egbib}
}

\end{document}